\documentclass[runningheads]{llncs}

\usepackage{eccv}

\usepackage{eccvabbrv}

\usepackage{graphicx}
\usepackage{caption}
\usepackage{subcaption}
\usepackage{svg}
\usepackage{booktabs}
\usepackage{multirow}

\usepackage[accsupp]{axessibility}  

\usepackage[pagebackref,breaklinks,colorlinks]{hyperref}

\usepackage{orcidlink}

\begin{document}

\title{Controllable Face Synthesis with Semantic Latent Diffusion Models} 

\titlerunning{Controllable Semantic Latent Diffusion Models}

\author{Alex Ergasti\orcidlink{0009-0005-8110-9714} \and
Claudio Ferrari\orcidlink{0000-0001-9465-6753} \and
Tomaso Fontanini\orcidlink{0000-0001-6595-4874}\and
Massimo Bertozzi\orcidlink{0000-0003-1463-5384}\and
Andrea Prati\orcidlink{0000-0002-1211-529X}}

\authorrunning{A.~Ergasti et al.}

\institute{University of Parma, Department of Architecture and Engineering, Parma, Italy
\url{https://github.com/ErgastiAlex/SCA-DM} \\
\email{\{alex.ergasti,claudio.ferrari2, tomaso.fontanini,\\ massimo.bertozzi,andrea.prati\}@unipr.it}}

\maketitle

\begin{abstract}
Semantic Image Synthesis (SIS) is among the most popular and effective techniques in the field of face generation and editing, thanks to its good generation quality and the versatility is brings along. Recent works attempted to go beyond the standard GAN-based framework, and started to explore Diffusion Models (DMs) for this task as these stand out with respect to GANs in terms of both quality and diversity. On the other hand, DMs lack in fine-grained controllability and reproducibility. To address that, in this paper we propose a SIS framework based on a novel Latent Diffusion Model architecture for human face generation and editing that is both able to reproduce and manipulate a real reference image and generate diversity-driven results. The proposed system utilizes both SPADE normalization and cross-attention layers to merge shape and style information and, by doing so, allows for a precise control over each of the semantic parts of the human face. This was not possible with previous methods in the state of the art. Finally, we performed an extensive set of experiments to prove that our model surpasses current state of the art, both qualitatively and quantitatively.
\end{abstract}

\section{Introduction}

\begin{figure}[tb]
    \centering
    \includegraphics[width=0.9\textwidth]{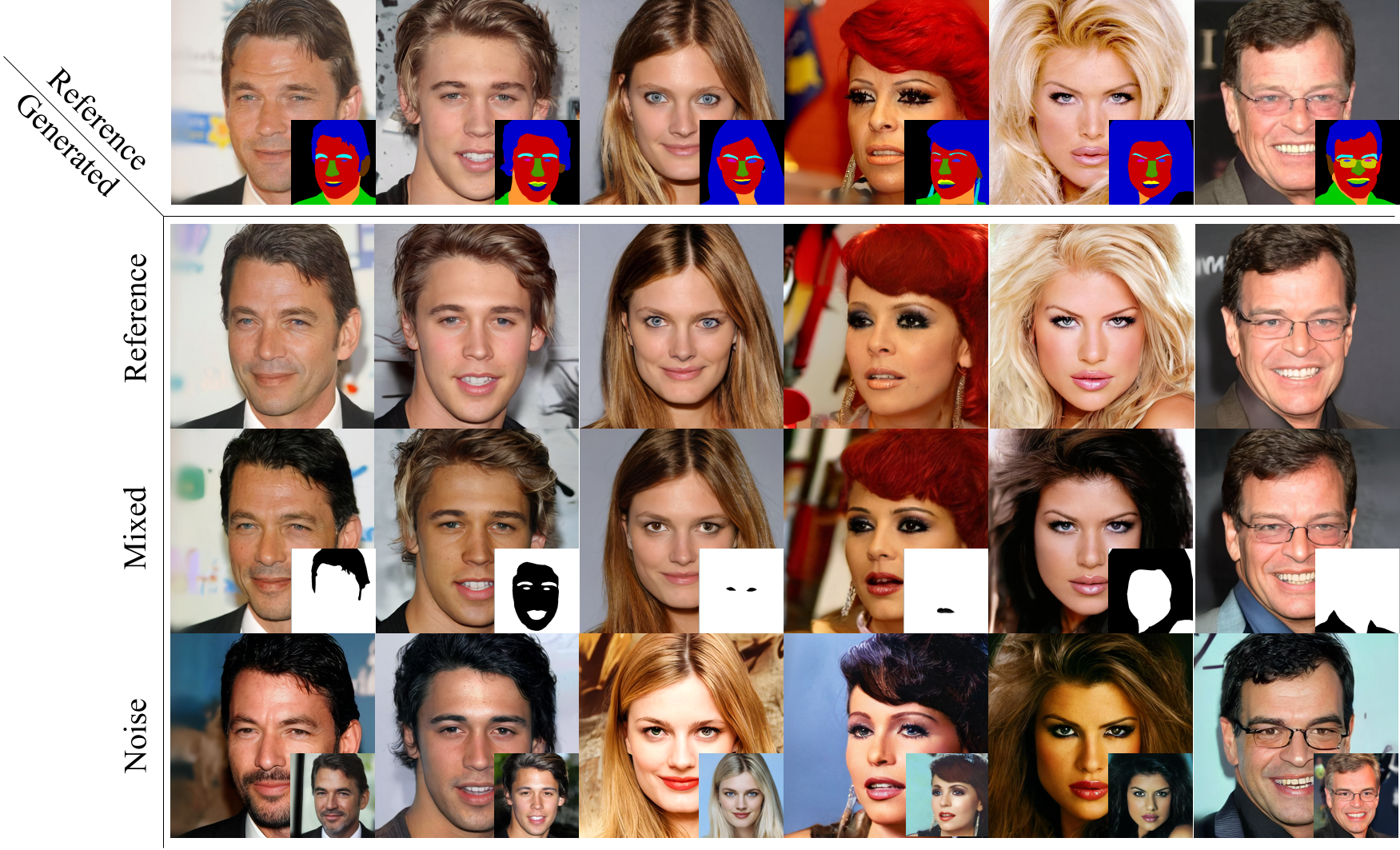}
    \caption{Our model can generate images in three ways: (a) Given a reference image, (b) mixing styles from a reference image and noise, (c) fully noise-based without reference.}
    \label{fig:results}
\end{figure}

\label{sec:intro}
Diffusion models (DMs) are the current state of the art in the majority of fields involving generative tasks. Since their introduction, they were applied to a large variety of scenarios, yet their most important application is in the field of image generation. In this domain, they were applied in several contexts, from text-to-image \cite{ldm} to semantic segmentation \cite{baranchuk2021label} and many more \cite{croitoru2023diffusion}, thanks to the excellent quality and variety of data that can be generated through them. However, even if results obtained with DMs are unquestionably impressive, they face two major limitations, that are \textit{(i)} the lack of controllability and reproducibility of the generated samples and \textit{(ii)} the missing ability of reproducing and editing real images. This means that it is both extremely difficult to have a precise and fine-grained control of what is generated, mostly at a local-level, and to manipulate only specific objects in a given image. In the attempt to overcome these problems, several solutions were proposed. Some methods focus on controlling the output of the generative process in terms of spatial arrangement or shapes, such as in ControlNet \cite{cnet}, Composer \cite{huang2023composer}, T2I \cite{t2i} or Semantic-Diffusion~\cite{SDM}. These architectures pursue the goal of controlling the layout and appearance of the generate images by using sketches, color palettes, depth or semantic maps as additional information. Other works instead aim to encode a set of reference images to reproduce an object and its appearance in order to generate new samples including such object, as in Textual Inversion (TI) \cite{textualInversion} or Dreambooth \cite{ruiz2023dreambooth}.

Scenarios where generative models are applicable is vast. Nowadays, pretty much any data can be generated. We chose to explore the domain of human faces for several reasons: first, generative face models raise severe safety and privacy related concerns, as effective models can be maliciously used to alter or change the identity of a given individual \textit{i.e.} DeepFakes. On the other hand, generative face models can be employed for biometric applications such as increasing the robustness of face recognition systems to adversarial attacks using synthetic data, achieving important privacy preserving properties. Finally, faces present a high degree of correlation across facial parts. A model is thus prone to reproduce biases that are commonly found in human face datasets \textit{e.g.} people with light colored eyes usually have blonde hair, which makes disentangling, controlling and generating diverse appearances for local face parts more challenging.
Nevertheless, in general, all the methods in the literature based on diffusion models are all designed to \textit{generate} new data, while trying to control the generation in some way. To the best of our knowledge, there is yet no convincing solution to exploit diffusion models for addressing the task of reconstructing and precisely manipulating \textit{real images}. Hence, in this paper we present a solution to augment diffusion models with the above capability.


More specifically, we cast the problem of face editing, manipulation and generation in a Semantic Image Synthesis (SIS) framework. This choice is motivated by the observation that SIS approaches demonstrated the most effective for the task of precise image manipulation. SIS methods address the task of generating photo-realistic images using their semantic mask as a condition to control the spatial layout. A semantic mask is an image in which each pixel represents a semantic class, and is pixel-wise aligned with the RGB reference. In the literature, the vast majority of SIS architectures make use of custom normalization layers to modulate the features activation with the information contained in the semantic mask. In particular, the most common paradigm is represented by SPADE \cite{spade} which introduced spatially-adaptive normalization layers. There exist several SIS methods with different goals: some focus of generating a random style for each semantic parts (noise-based) \cite{tan2021diverse, richardson2021encoding}, while others are trained to extract specific styles from a reference image and map them to each semantic region (reference-based) \cite{sean, tan2021diverse}. Until recently, the majority of SIS models were based on Generative Adversarial Networks. With the surge of diffusion models, the research efforts started moving towards them. In particular, the most relevant diffusion SIS architecture is Semantic Diffusion Model (SDM) \cite{SDM} which fuses the powerful generation capability of DM with SPADE normalization layers in order to precisely control the shape of the generated samples. Still, SDM is fully noise-based and cannot reproduce a set of specific styles during generation. 





To overcome this limitation, in this paper we propose a novel diffusion-based SIS model, named Semantic Class-Adaptive Diffusion Model (SCA-DM), that is able to both generate diverse samples conditioned on a semantic mask but, at the same time, can also be used to extract precise styles from any reference image with the specific goal of accurate human face editing. Thus, our goal is to design and investigate the first diffusion-based architecture that encapsulates both the above features. We do so by adding a style reference image in addition to the semantic mask as condition to guide the diffusion process. To achieve that, we trained from scratch a Latent Diffusion Model (LDM), conditioned with both the semantic mask and styles extracted from the reference image by means of a dedicated style encoder. Conditioning a diffusion model can be achieved in multiple ways, such as via Cross Attention, concatenation or, as in our specific case, with SPADE normalization layers. Since our proposed architecture needs to merge both the information provided by the mask and the reference image to be edited, we chose to use a variation of the Cross Attention to control the style condition, and SPADE to condition the spatial layout with the semantic mask. 


We will show that the proposed solution enables both accurate face generation and editing of real images, a property that is still under-explored with diffusion models, and showcase its advantages with respect to prior GAN-based methods and the recent SDM~\cite{SDM}. In sum, the contributions of this work are:
\begin{itemize}
    \item We developed a novel LDM architecture for SIS that is both able to generate diverse samples and exactly reproduce a given style extracted from a reference image, opening the way to precise human face editing.
    \item We propose and explore the combination of SPADE normalization layers and cross-attention to fuse layout and style information together, as opposed to prior works that only use SPADE. Additionally, we designed a modified mask-conditioned cross-attention layer in order to improve the disentanglement of different facial parts.
    \item We provide an extensive set of qualitative and quantitative evaluation to showcase the proposed model capability and its superiority w.r.t. to current state-of-the-art solutions.
\end{itemize}

\section{Related Work}\label{sec:related}

\noindent\textbf{Diffusion Models.}
Diffusion models are currently the state of the art of generative models, firstly proposed by \textit{Ho et Al.} \cite{ho2020denoising} as Denoising Diffusion Probabilistic Models (DDPM). Numerous studies have focused on enhancing diffusion models to mitigate their primary drawbacks, namely prolonged training and inference time. To address the former, the Latent Diffusion Model (LDM) \cite{ldm} was proposed introducing an encoder/decoder structure to reduce the dimensionality of the space in which the diffusion model operates. By enabling the diffusion model to function within a latent space, researchers have successfully reduced the required training time.
To tackle the latter, a new sampling technique called DDIM \cite{song2022denoising} has been proposed. It allows to sample from a DM without any re-training requirements, reducing in this way the steps required by paying just a little of image quality, speeding up the process even up to a factor of $20$. Moreover, the advancement of diffusion models has opened up for various works aimed at enhancing their capabilities. For instance, works like Control Net \cite{cnet}, Composer \cite{huang2023composer} or T2I-Adapter \cite{t2i} represent notable expansions of LDM. All of these approaches were proposed with the common objective of enhancing the control and precision of image generation in diffusion models, incorporating different condition mechanism, such as semantic masks or scribbles.

Other works based on pretrained LDM focus on adding new information inside the model, by teaching a new word $S^*$, as in Textual Inversion \cite{textualInversion} and Dreambooth \cite{ruiz2023dreambooth}. Nevertheless, both methods require a fine-tuning without being one shot and are not tailored to exactly reproduce human faces but rather subjects for which inconsistencies are less spottable by a human eye. 
Other disadvantages of these techniques are that they do not allow a precise style mixing, delegating all the generation variety to the chosen text, thus without any possibility to a fine grained control over the the generated samples.

\noindent\textbf{Semantic Image Synthesis.}
The current landscape of semantic image synthesis can be broadly divided into two main categories: GAN-based and Diffusion-based approaches. Within the GAN-based category, further subdivision is possible based on whether the models are reference-based or not. Under the former category, numerous GAN models exist, many of which share a similar structure. 

In the current state of the art, SEAN \cite{sean} is a notable model that utilizes SPatial Adaptive DE-normalization layers (SPADE, \cite{spade}) to inject both semantic masks and style embeddings into the generator. However, it lacks the ability to generate diverse images; given the same input, it always produces the same image.
Another noteworthy model is CLADE (CLass Adaptive DE-normalization) \cite{clade}, which has the capability to generate images both with and without an RGB reference image. This flexibility enhances its applicability in various scenarios. Additionally, INADE \cite{inade} is another significant model in the landscape of semantic image synthesis. Unlike SEAN, INADE focuses on producing images using instance style rather than class style. This allows for conditioning each instance in the image individually, leading to more precise and nuanced results. Next, FrankenMask \cite{fontanini2023frankenmask} focused on the automatic manipulation of the semantic mask. Another remarkable model is SemanticStyleGAN \cite{shi2022semanticstylegan}, which innovatively expands upon StyleGAN \cite{karras2019stylebased}. Unlike other cited approaches, SemanticStyleGAN does not directly utilize a given semantic mask during the generation process. Instead, it splits the style embedding with the semantic mask embedding information. This novel approach enables the model, with the aid of GAN inversion \cite{zhu2020domain} techniques, to generate a face given a semantic mask. Finally, Tarollo \textit{et al.} \cite{tarollo2024adversarial} employed a SIS model to perform adversarial attacks on face recognition systems, proving the broader range of applications of these models.

One of the most renowned and accessible diffusion models for semantic image synthesis is the Semantic Diffusion Model (SDM) \cite{SDM}. Unlike other models, SDM is conditioned solely by the semantic mask without any reference image. Consequently, it excels in generating diverse images, but without the option to guide the generation through a reference image.
Other notable models in this domain include Control Net \cite{cnet}, with every other similar works as Composer \cite{huang2023composer} or T2I \cite{t2i}, and Collaborative Diffusion \cite{cdiffusion}. Control Net (and similar) incorporates semantic masks using UNet injection, providing additional control over the synthesis process. On the other hand, Collaborative Diffusion utilizes multiple diffusion models that collaborate with each other to produce high-quality face images, showcasing the potential for collaborative approaches in SIS.

The goal of our research is to propose a Latent Diffusion Model (LDM) that is able to produce high quality and diverse samples, which shape is controlled by semantic masks while the styles can be both extracted from a reference image or generated from scratch, as shown by \cref{fig:results}. Our system is capable of maintaining the overall coherence and allowing controllability of the generated images better than current state of the art in every of its different configurations.


\section{Proposed Architecture}

\begin{figure}[tb]
    \centering
        \includegraphics[width=\textwidth]{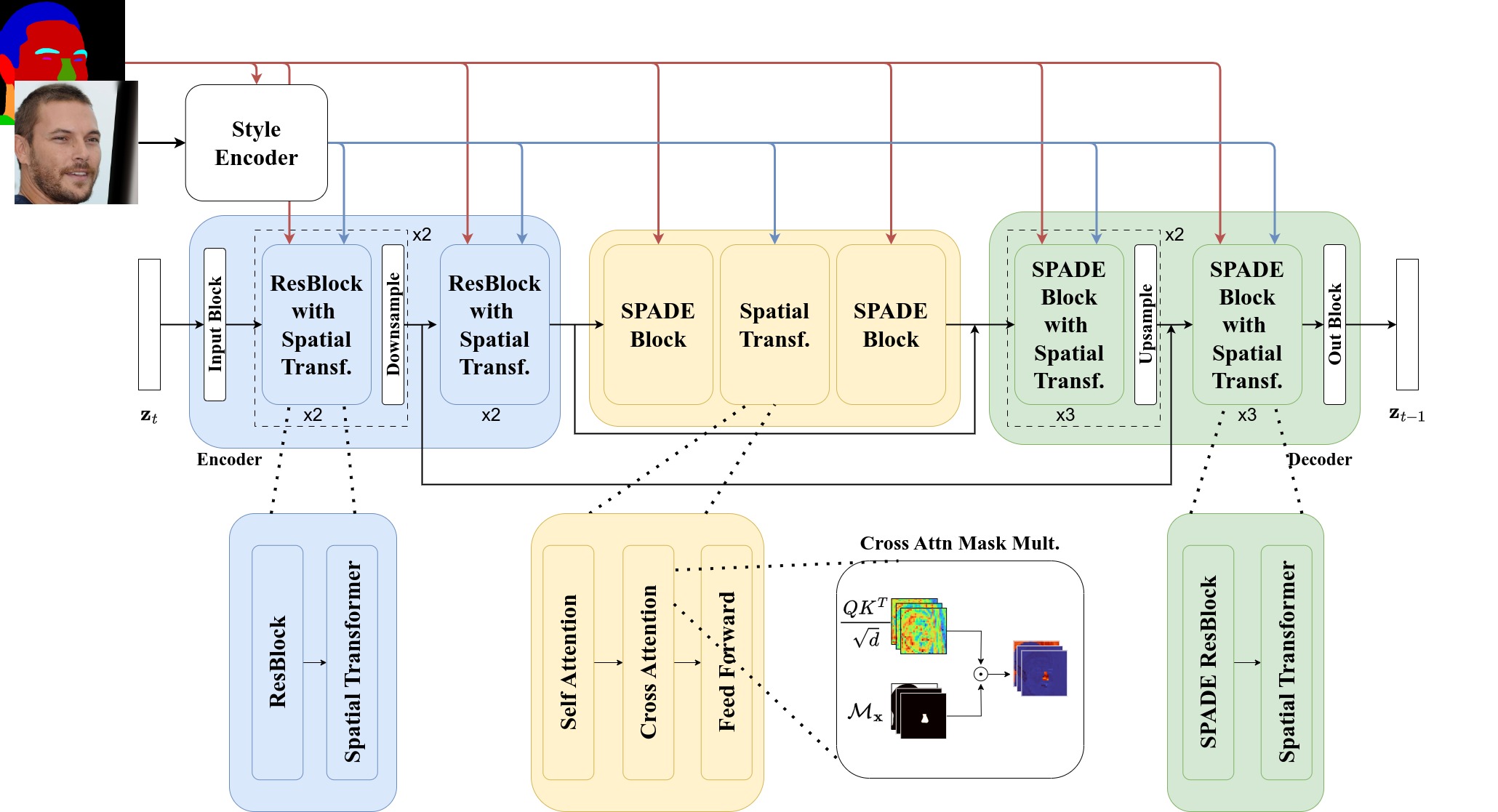}
        \caption{Model architecture. The encoder part of the UNet uses only standard Resnet Block with SpatialTransformer to guide the diffusion process with the style embedding obtained from $\mathcal{E}_s$. The middle block and the decoder part use SPADEResBlock, as in SDM, to encapsulate the semantic mask info.
        The Mask attention mechanism is applied inside the SpatialTransformer on the Cross Attention Map.}
        \label{fig:architecture}
\end{figure}

We propose SCA-DM (Semantic Class-Adaptive Diffusion Model), a Latent Diffusion Model (LDM \cite{ldm}) for semantic face image synthesis. Our model is composed of: a pretrained VQGAN, a custom Diffusion Model which integrates SPADE and cross-attention layers, and a Multi-Resolution Style Encoder to extract the style given an RGB image and the corresponding semantic style mask. Our custom diffusion model has SPADE ResBlock in the middle block and in the decoder block, inspired by SDM \cite{SDM}, and it uses Spatial Transformer to inject the Style Embedding inside the UNet architecture, as shown by \cref{fig:architecture}.

During training, the model will learn to merge shape and style information together in order to faithfully reproduce any given subject. 

\subsection{Latent Diffusion Model}
Diffusion models are probabilistic models designed to gradually remove noise from a normally distributed sample $\mathbf{z}$. The process that gradually adds Gaussian noise with fixed and scheduled variance $\beta_1,\dots,\beta_T$ to the data $\mathbf{x}_0$ is called forward process and it is described by:
\begin{equation}
    \begin{split}
         q\left(\mathbf{x}_{T}|\mathbf{x}_{0}\right) &=\prod_{t=1}^Tq\left(\mathbf{x}_t|\mathbf{x}_{t-1}\right)\\
         q\left(\mathbf{x}_t|\mathbf{x}_{t-1}\right) &=\mathcal{N}\left(\mathbf{x}_t;\sqrt{1-\beta_t}\mathbf{x}_{t-1};\beta_t\mathbf{I}\right)
    \end{split}       
\end{equation}
The model learns a data distribution $p(\mathbf{x})$, which corresponds to learning the reverse process of a fixed Markov chain of length $T$. The most successful models \cite{dmbeatgan,dm} rely on a reconstruction loss of the image $\mathbf{x}_{t-1}$ given $\mathbf{x}_{t}$, which is indeed a reweighted variant of the variation lower bound on $p(\mathbf{x})$. In our case, the training loss is defined as follows:

\begin{equation}
    \mathcal{L}_{\mathrm{DM}}=\mathbb{E}_{\mathbf{x},\epsilon\sim\mathcal{N}(0,1),t}\left[\left\lVert\epsilon-\epsilon_\theta\left(\mathbf{x_t,t,\mathcal{E}_s\left(x\right),\mathcal{M}_x}\right)\right\rVert_2^2\right]
\end{equation}
The model architecture $\epsilon_\theta$ is depicted in \cref{fig:architecture}. It is conditioned with time $t$, semantic mask $\mathcal{M}_\mathbf{x}$ and the style embedding $\mathcal{E}_s(x)$ of the given style image $x$. By doing so, the proposed system will link the shape information of the generated sample to $\mathcal{M}_\mathbf{x}$ and map the extracted semantic styles to each semantic region. Since, as said before, the forward process is fixed, during training it is possible to obtain $\mathbf{x}_t$ directly for each uniform sampled $t$ via:
\begin{equation}
    \mathbf{x}_t=\sqrt{\overline\alpha_t}\mathbf{x}_0+\sqrt{1-\overline\alpha_t}\epsilon, \quad \mathrm{where}\: \overline{\alpha}_t=\prod_{i=1}^t\alpha_i, \:\alpha_i=1-\beta_i.
\end{equation}

Our proposed model is a Latent Diffusion Model, which means it works on latent space and not directly on the image space. Because of this, we use a pretrained VQGAN to encode and decode the image from and to the latent space. Apart from the pretrained VQGAN, both the model $\epsilon_\theta$ and the style encoder $\mathcal{E}_s$ are trained together using the aforementioned loss. It is also possible to encourage the model to place greater reliance on conditioning, as demonstrated by Ho et al. \cite{ho2021classifierfree}. This can be achieved by computing and combining the estimated noises ($\epsilon_\theta$) with and without conditioning, obtaining a new noise $\overline\epsilon$. This approach involves passing an empty conditioning through the network to compute the estimated noise, thereby obtaining a refined estimation that incorporates the influence of conditioning terms $\mathcal{E}_s(x)$ and $\mathcal{M}_\mathbf{x}$:

\begin{equation}
    \begin{split}
    \overline{\epsilon}(\mathbf{x_t,t,\mathcal{E}_s(x),\mathcal{M}_x}) & = \epsilon_\theta(\mathbf{x_t,t,\mathcal{E}_s(x),\mathcal{M}_x})+ \\& +s(\epsilon_\theta(\mathbf{x_t,t,\mathcal{E}_s(x),\mathcal{M}_x})-\epsilon_\theta(\mathbf{x_t,t,\mathcal{E}_s(\emptyset),\emptyset}))      
    \end{split}
    \label{eq:uc}
\end{equation}

\noindent where $s$ controls the intensity of the combination.

\subsection{Multi-Resolution Style Encoder}
The Multi-Resolution Style Encoder $\mathcal{E}_s$ takes as input an RGB image $x\in \mathbb{R}^{3\times H\times W}$ and a corresponding semantic mask image $\mathcal{M}_x\in \mathbb{N}^{C\times H\times W}$. For each of the $L$ convolutional layers in the style encoder, we extract style features specific to each of the $C$ semantic classes. Specifically, at the i-th layer, with feature maps of size $\mathbb{R}^{D_i\times H_i\times W_i}$, we split the $D_i$ channels into $C$ groups, yielding a feature map $F_{i,j}\in \mathbb{R}^{D_i/C\times H_i \times W_i}$ for each semantic class $j$. We then apply element-wise multiplication with the corresponding mask channel, followed by average pooling ($AP$), to obtain the style feature $S_{i,j}$:

\begin{equation}
    S_{i,j} = AP\left(F_{i,j}\cdot \mathcal{M}_j\right)
\end{equation}

\noindent
Subsequently, we reshape each $S_{i,j}$ to have $D$ channels using a $1\times 1$ convolution. This process is repeated for each layer $i=1,\dots,L$, resulting in a multiscale style embedding $S\in \mathbb{R}^{C\times L \cdot D}$, where $D$ is the size of the embedding at any layer.

\subsection{Mask-conditioned Cross-Attention}

During training, our model makes use of cross-attention layers in order to map the style codes in the corresponding semantic regions of the images. More in detail, cross-attention layers are defined as follows:

\begin{equation}
    \mathcal{L}_{CA}(Q,K,V) = \mathcal{S}\left(\frac{QK^T}{\sqrt{d}}\right)V
    \label{eq:att_map}
\end{equation}

\noindent where $\mathcal{S}$ is the Softmax function. Additionally, differently than self-attention that combines embeddings from a single source, in cross-attention embeddings come from two separate sources. In particular, in our case, $Q$ is obtained from the projection of the features of previous convolutional layers $\mathcal{F}_i$, while $K$ and $V$ are obtained from the projection of the style codes $\mathcal{E}_s(x)$. More in detail:

\begin{equation}
    Q=W_Q^{(i)} \cdot \mathcal{F}_i, 
    K = W_K^{(i)} \cdot \mathcal{E}_s(x,\mathcal{M}_x), 
    V = W_V^{(i)} \cdot \mathcal{E}_s(x, \mathcal{M}_x)
\end{equation}

In the proposed architecture we modify the cross-attention layers in order to force disentanglement between each class embedding extracted by the Style Encoder. More in detail, we multiply in each Cross Attention the given attention map with the semantic mask (as shown in \cref{fig:architecture}). By doing this, a single style embedding can only modify a very specific part of the image, therefore encouraging the style encoder to extract only local information, delegating to self-attention the image coherency. A cross-attention layer then becomes:

\begin{equation}
    \mathcal{L}_{CA}(Q,K,V) = \mathcal{S}\left(\frac{QK^T}{\sqrt{d}}\mathcal{M}_x\right)V
    \label{eq:att_map}
\end{equation}
By multiplying the mask \textit{before} applying the Softmax function we make sure that the numerical coherence is maintained. 

\subsection{SPADE ResBlock}
In our architecture, the Multi Style Encoder $\mathcal{E}_s$ is responsible for the style embedding and cross attention layers inject the extracted style in the diffusion process. In order to inject also the semantic mask $\mathcal{M}_x$, we substitute the basic ResBlock of U-Net backbone with SPADE ResBlock ~\cite{spade, SDM} (Fig. ~\ref{fig:architecture}). The SPADE ResBlock, an extension of the ResNet block ~\cite{he2016deep}, is equipped with SPADE layers that take the semantic image mask ($\mathcal{M}_x$) as a conditioning input. SPADE layers work by generating two feature maps, $\gamma$ and $\beta$, which are subsequently used to scale and shift the features inside the ResBlock. By incorporating these blocks, we enable the diffusion model to be conditioned on the semantic mask.

\section{Experiments}
\noindent\textbf{Training Details.} We train our model on a NVIDIA 4090 gpu with a learning rate of $2.0e-06$. During training, $50\%$ of the images to the style encoder were set to zeros; this lets the model learn to produce random images without any reference, and improved the capability of generating conditioned images manipulating the scale $s$ in \cref{eq:uc}. For all the tests, we empirically set $s=1.2$.

\noindent\textbf{Dataset.} we train our model on CelebAMask-HQ \cite{celeba} which is composed by 30k human face images paired with the corresponding 19-channel semantic mask. The dataset is splitted as follows: 28k images for training and 2k for testing.

\noindent\textbf{Metrics.}
To evaluate our method, we employ the current state-of-the-art evaluation metrics of SIS methods. In particular, we use Frechet Inception Distance \cite{heusel2017gans} (FID) to estimate the generation quality compared with the test dataset. Furthermore, we evaluate the similarity between the given semantic mask and the semantic mask parsed from the generated images using mean Intersection-over-Union (mIOU) and pixel accuracy. We use a pretrained FaceParsing model \cite{faceparsing} to generate the masks from the fake samples. Finally, to validate how closely the model is able to reconstruct the reference human face, we calculate Structural Similarity Index Measure (SSIM). We test our model in the following experimental settings: reconstruction (reference-based), total style swap, partial style swap and diversity (noise-based).

\subsection{Results on reconstruction}

\begin{figure}[tb]
    \centering
    \includegraphics[width=0.85\textwidth]{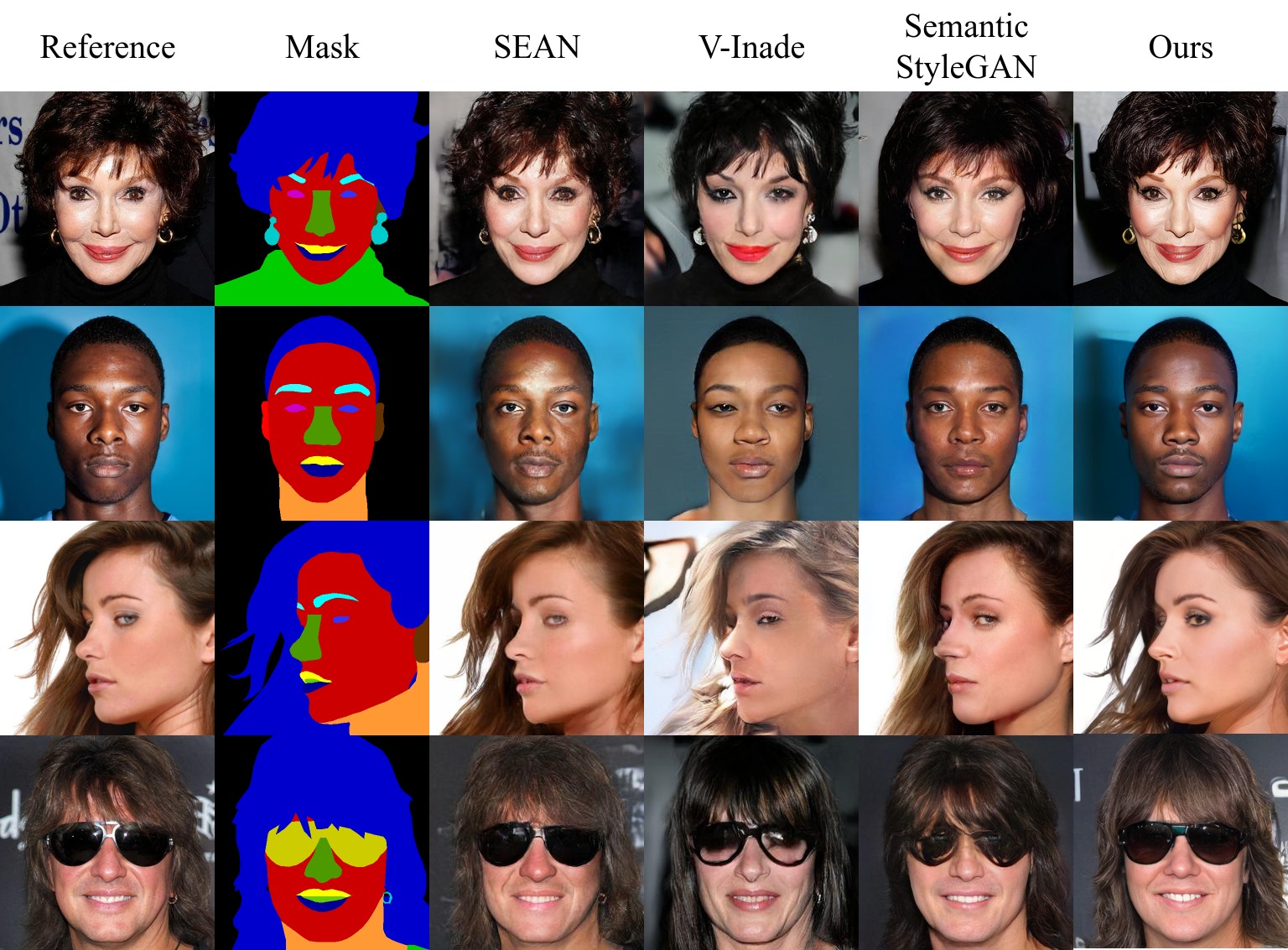}
    \caption{Image reconstruction comparison between the state of the art and our model.}
    \label{fig:rec}
\end{figure}

\begin{table}
\centering
\begin{tabular}{c|cccc}
\toprule
\multirow{2}{*}{\textbf{Architecture}} &  \multicolumn{4}{c}{\textbf{CelebA-HQ}}\\
  &
  \textbf{SSIM} $\uparrow$ & \textbf{FID} $\downarrow$ & \textbf{mIOU} $\uparrow$ & \textbf{Acc.} $\uparrow$ \\
\midrule
MaskGAN \cite{celeba} & 0.51 & 59.91 & 76.3 & 87.8 \\
SEAN \cite{sean} & 0.49 & 20.82 & \textbf{82.4} & \textbf{95.0}  \\
V-INADE \cite{inade} & 0.29 & 17.49 & 78.0 & 93.5 \\
SemanticStyleGAN \cite{shi2022semanticstylegan}& 0.52  & 26.83 & 69.8 & 90.2 \\
\midrule
\midrule
\textbf{Ours}  & \textbf{0.54} &  \textbf{16.85} & 81.78 & 94.67 \\
\bottomrule
\end{tabular}
\caption{Comparison with the state of the art of reconstruction models in terms of SSIM, FID, mIOU and segmentation accuracy.}
\label{tbl:recon_comparison}
\end{table}

As a first experiment, we tested the capability of our model to faithfully reconstruct any human face starting from the corresponding semantic mask and styles. A qualitative evaluation of this task can be seen in Fig. \ref{fig:rec} where different reconstruction results produced by several state-of-the-art methods are compared to the results generated by our model. Our model can precisely reconstruct the input image even in challenging setting (Fig. \ref{fig:rec}, bottom rows). 

Then, we performed an extensive quantitative evaluation, as shown in \cref{tbl:recon_comparison}.  Our model achieves better FID compared to state-of-the-art models. On the other side, we have slightly worst performance in terms of mIOU and Accuracy even if we still position as second-best. This is likely caused by the fact that the proposed diffusion model works in the latent space, using an VQGAN encoder/decoder architecture to go from and to pixel space leading to some misalignment between mask and generated image. Finally, we calculate SSIM to verify the similarity between the generated samples and the reference image and we obtained the best result overall. In this section, a comparison with SDM was not possible given its inability to perform reference-based inference.

\subsection{Results on total style swap}
\begin{figure}
    \centering
    \includegraphics[width=0.75\textwidth]{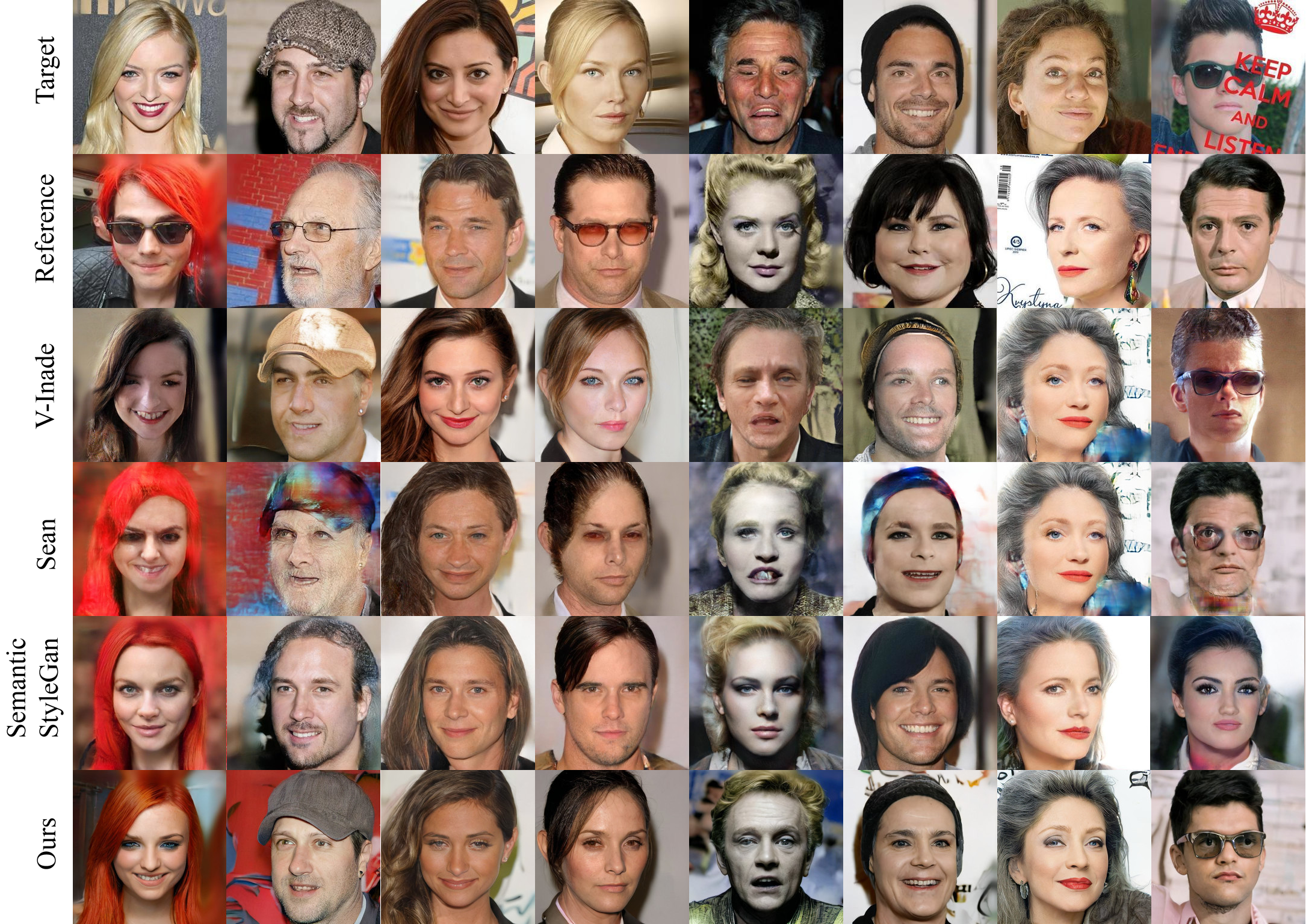}
    \caption{Style transfer comparison between different methods and our model. The style of the reference image is applied to the target image. The overall consistency in style swap is far better compared to state-of-the-art methods.}
    \label{fig:style_swap}
\end{figure}


We conducted a comparative analysis to assess our model's ability to generate images using the style of another image (total style swap). The quality of the generated images is shown in \cref{fig:style_swap} where all models capable of style transfer were presented. Our system demonstrates superior overall consistency and produces images of higher quality compared to the state of the art. Additionally, thanks to our training approach, the model can generate a random style for elements not present in the reference image, as showcased in the second column where the hat is accurately generated despite its absence in the reference. Similarly, in the first column, the model generates meaningful eyes without explicit reference, highlighting its versatility.

Furthermore, to provide additional evidence of our model capabilities both in terms of flexibility and robustness, we showcase in \cref{fig:partial_interpolation} (bottom two rows) the interpolation of a full style embedding.

\subsection{Results on partial style swap}
\begin{figure}[tb]
    \centering
    \includegraphics[width=0.65\textwidth]{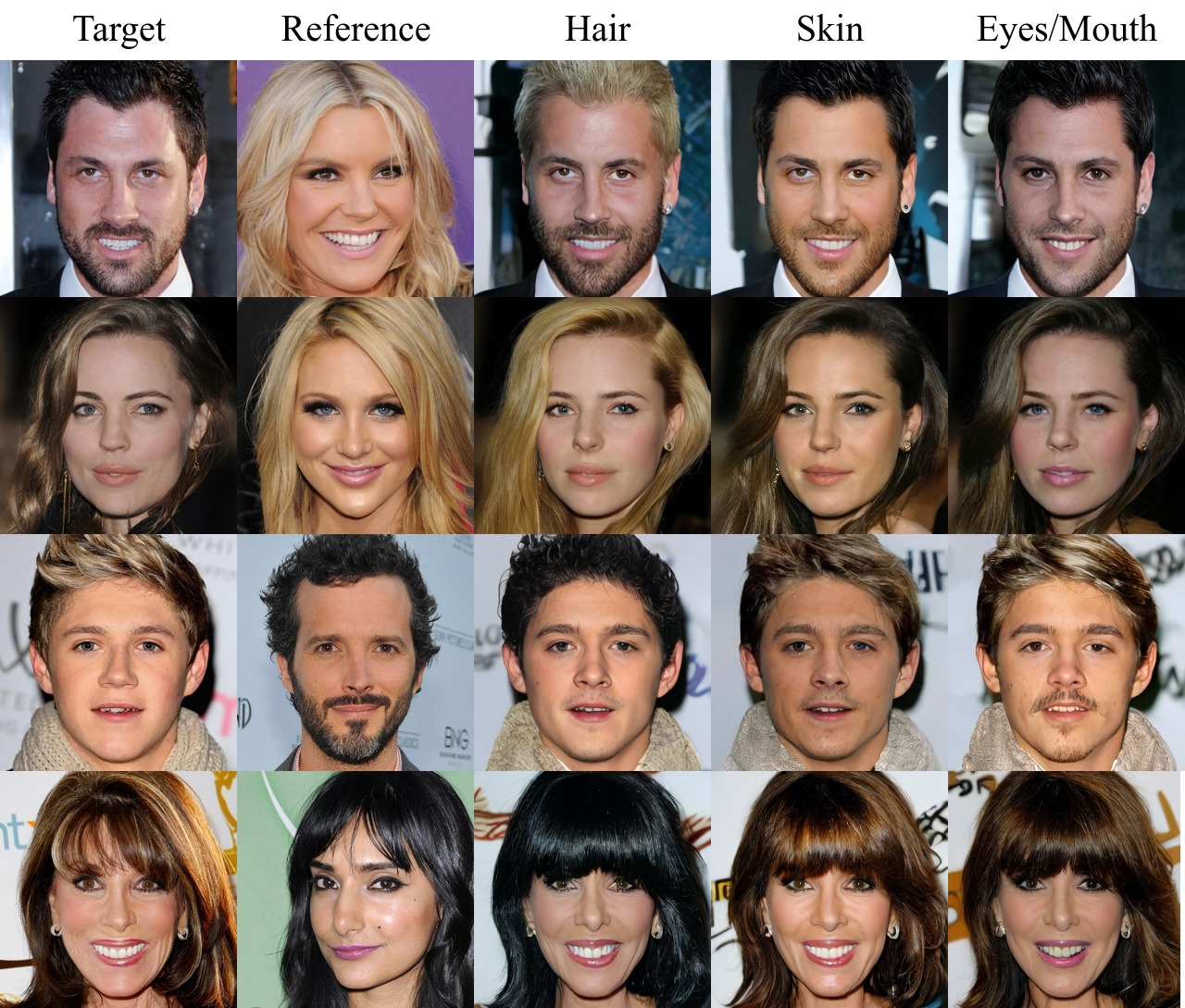}
    \caption{Style transfer of single face features. Our model can successfully swap single style part with high image coherence.}
    \label{fig:single_style_swap}
\end{figure}
\begin{figure}[tb]
    \centering
    \includegraphics[width=0.9\textwidth]{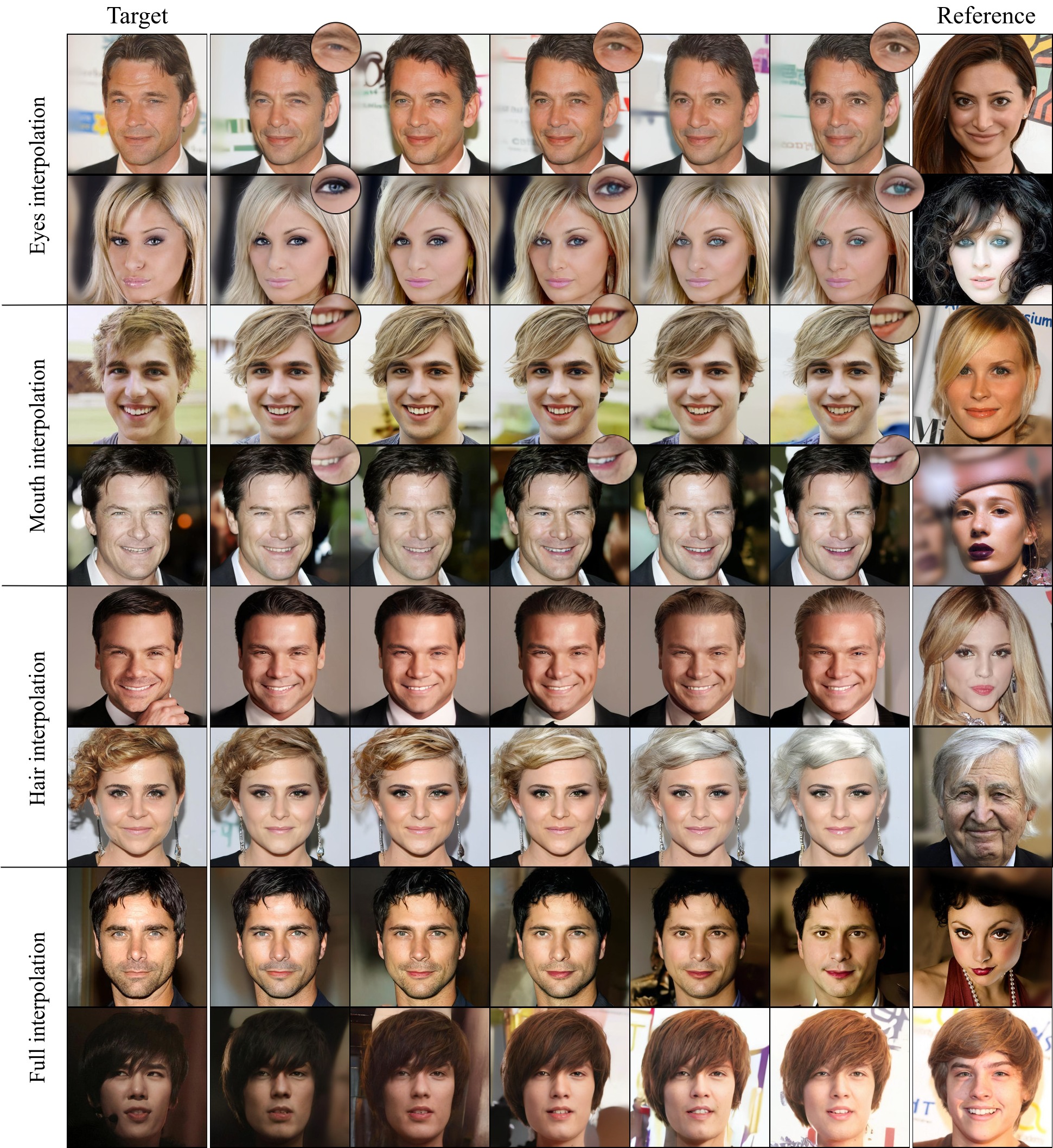}
    \caption{Interpolation of eyes, mouth, hair style and full style going from full target (left) to full reference (right). Some details are highlighted for a clear observation of changes.}
    \label{fig:partial_interpolation}
\end{figure}
\begin{figure}[tb]
    \centering
    \includegraphics[width=0.60\textwidth]{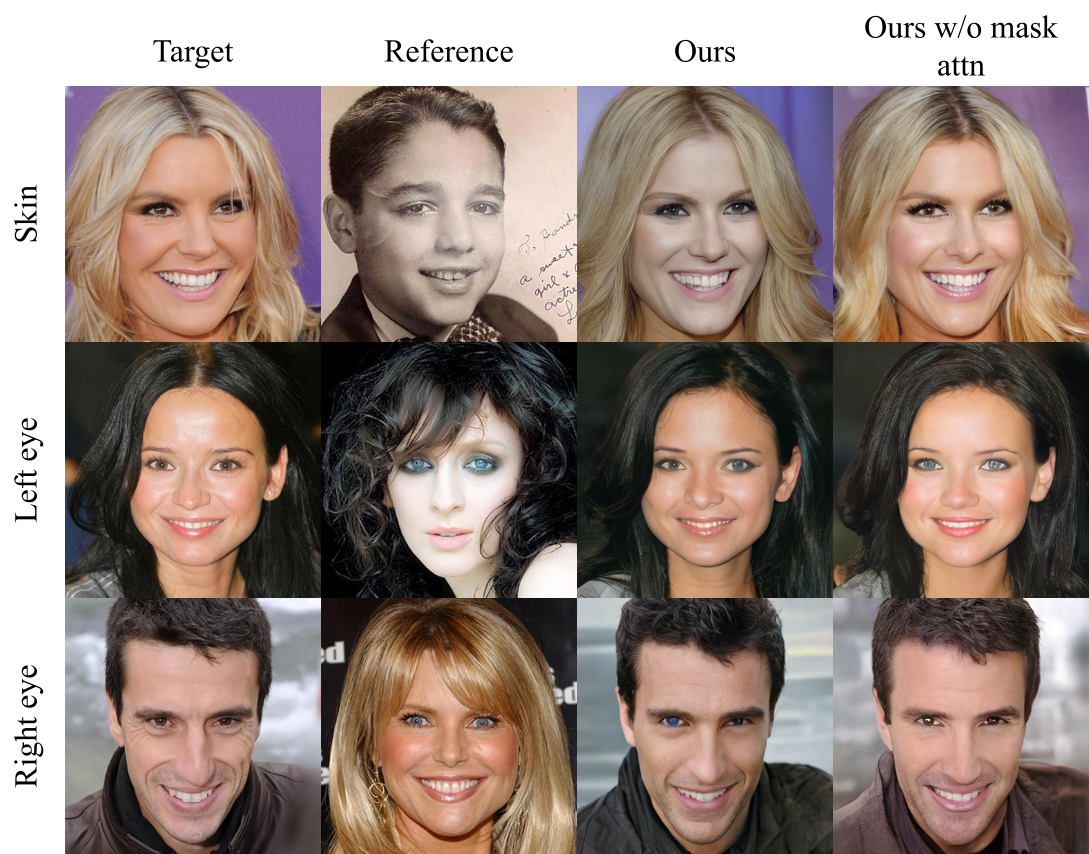}
    \caption{Ablation study which compares the model with and without mask attention. The model without mask attention has more entanglement between each style feature.}
    \label{fig:ablation}
\end{figure}
In this section, our objective is to demonstrate our model's ability to selectively alter only a portion of the style in the target image, as illustrated in \cref{fig:single_style_swap}.  Our model successfully swaps a subset of styles in the generated image, preserving the integrity of the other elements. 

This test also serves the purpose of demonstrating the efficacy of the masked attention that is crucial for this task. Indeed, without it, there would be too much entanglement between each style part and the model would not be able to generate images with mixed style, as evidenced in \cref{fig:ablation}. As expected, the model without the masked attention is not able to partially edit the style, \textit{e.g.} the skin tone does not change in the first row. In the last two rows, we tried to change the style of a single eye. Because the eyes style is entangled (the eye color of every face in the dataset is the same for both eyes), the model without any mask attention is not able to perform this local editing, but the style is applied to both.

Further evidence of our model proficiency in partial style swapping can be found in \cref{fig:partial_interpolation}, where we showcase its ability to interpolate partial feature styles from full target to full reference.

\subsection{Results on noise-based Generation}
\begin{figure}[tb]
    \centering
    \includegraphics[width=0.55\textwidth]{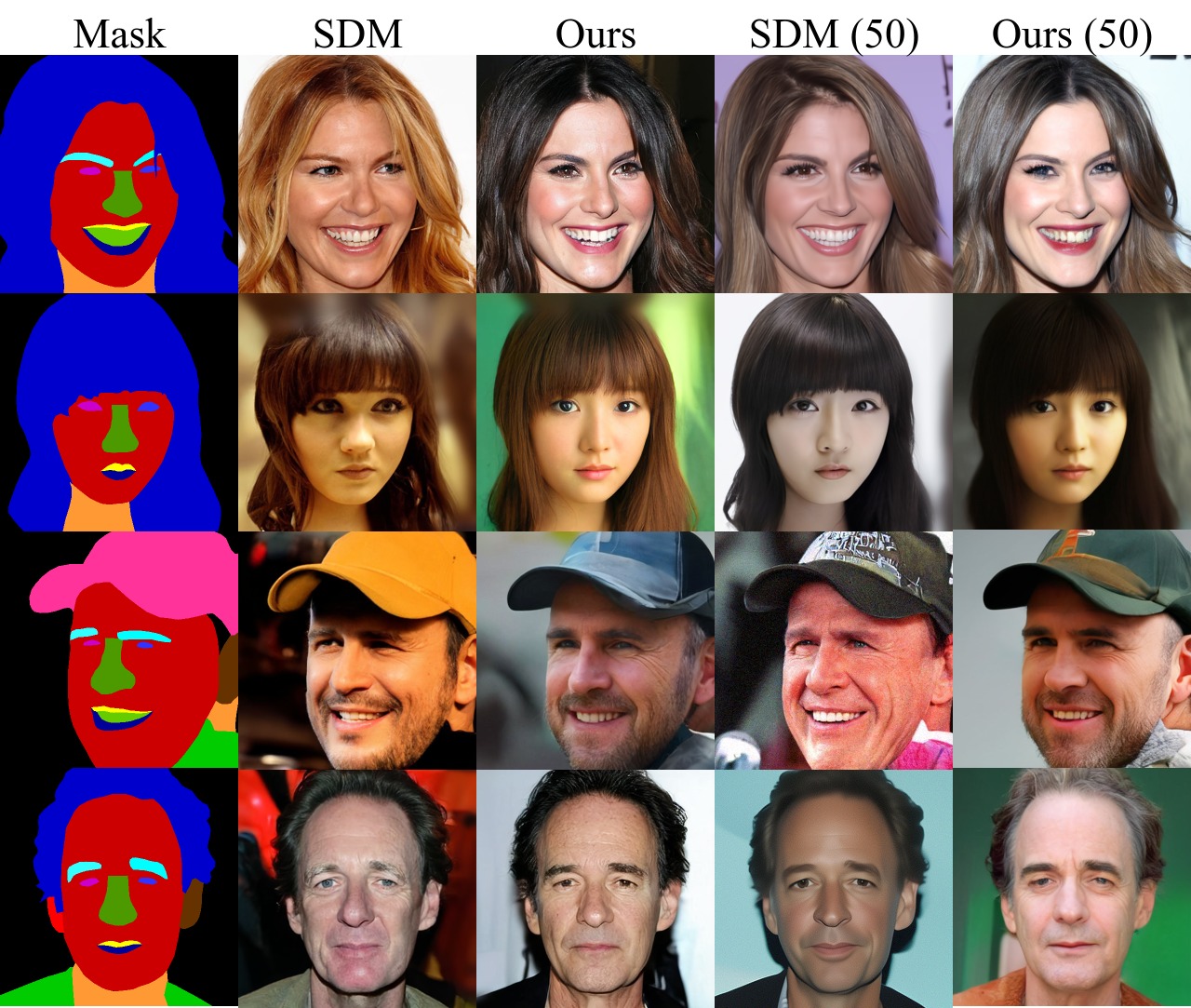}
    \includegraphics[width=0.4\textwidth]{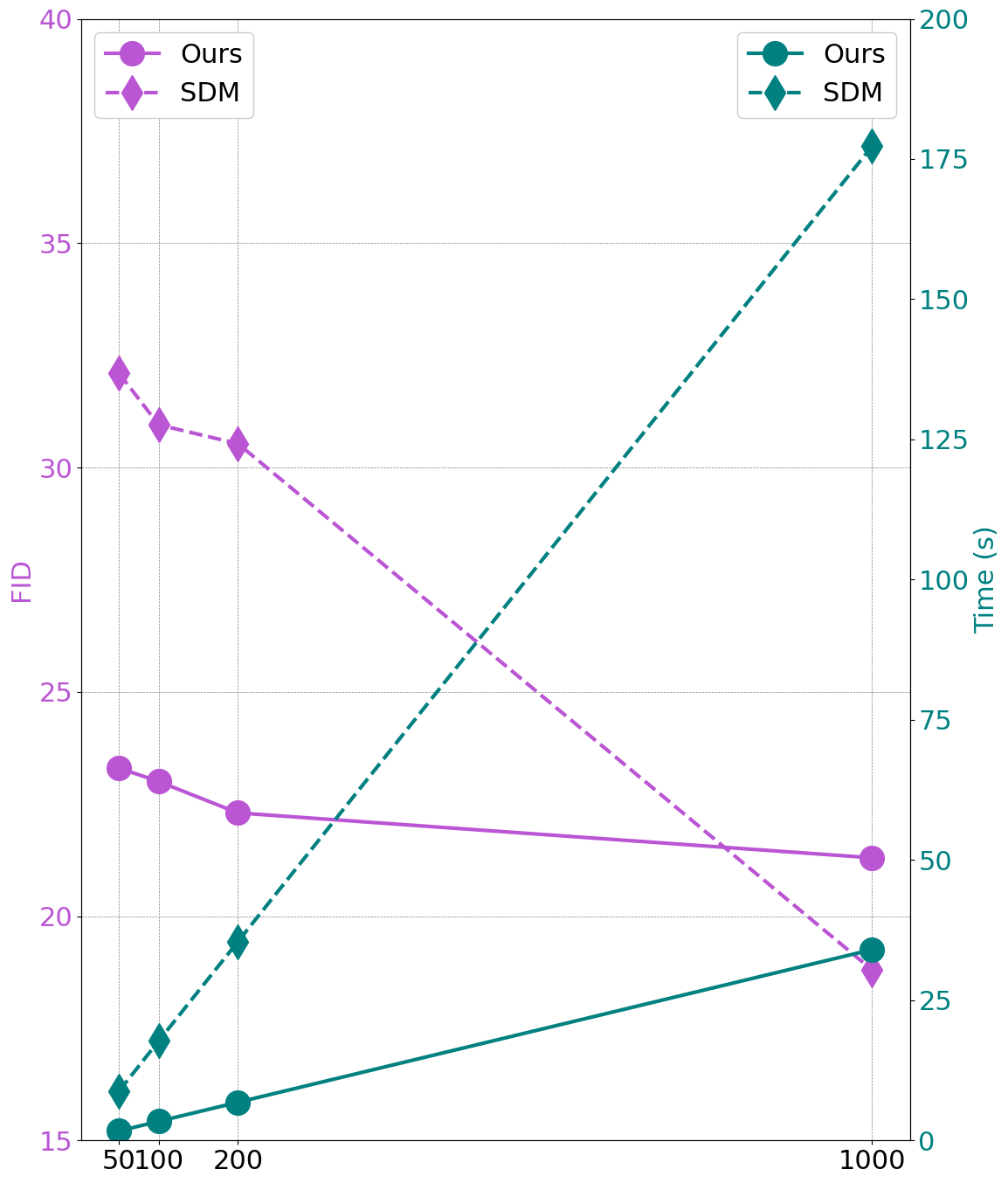}
    \caption{On the left: comparison of capability of generating images with random style between SDM and our model with 1000 or 50 diffusion steps. On the right: quantitative evaluation of FID and inference time for different numbers of diffusion steps.}
    \label{fig:sdm_vs_ours}  
\end{figure}

Our model is also able to generate images without any reference, by passing an all zero image to the style encoder. This allows our model to be compared also to solutions, like Semantic Diffusion Model (SDM), that can only generate diverse results but are unable to exactly reproduce a specific human face.

We reported both a quantitative and a qualitative evaluation in \cref{fig:sdm_vs_ours}. Starting from the former (on the right), a comparison between our model and SDM when using different diffusion steps for generation is presented. In particular, FID and inference time (in seconds) were calculated for 50, 100, 200 and 1000 diffusion steps. Indeed, the FID score of our model outperforms SDM in every  configuration except the last one. We argue our model performs better  with fewer diffusion steps due to the fact that it works in the latent space and, for this reason, can recover, thanks to the VQGAN, details that would have been lost with a small number of diffusion steps. This is crucial since, looking at the inference time, performing inference with 1000 steps is often unpractical, especially for SDM which overall takes 5$\times$ more time than our method to perform a single inference step. Finally, we calculated LPIPS to measure the diversity in the generated results, with an higher value corresponding to better diversity. Our model scores an LPIPS score of 0.33 against 0.42 of SDM. This was expected as adding reconstruction capability to a model can slightly hinder its generation of diverse sample due to some overfitting, \textit{e.g.} the model learns to associate semantic mask shapes to specific styles. 

Finally, we argue that our results are visually on par or even better than those produced by SDM expecially when a small number of diffusion steps is employed during inference. Indeed, this can be seen in \cref{fig:sdm_vs_ours} on the left.


\section{Conclusion}
In this paper, we proposed a novel SIS model for precise semantic editing and human faces synthesis. Our system is based on LDM \cite{ldm} and introduces in the decoder of the U-Net architecture a series of SPADE layers to condition the generation with the semantic masks. Additionally, masked cross-attention layers are employed to map styles extracted from a reference image to the semantic region of the masks without causing entanglement. With the combination of SPADE and cross-attention layers our model is able to both generate diverse results and to faithfully reproduce any given face. Regarding diversity, our model suffers from some overfitting during training and therefore sometimes struggle to generate samples completely different than the original RGB image. Additionally, the system allows for local and global style transfer, as well as style interpolation. Indeed, with current diffusion models conditioned with semantic information like SDM \cite{SDM} this was not possible, making our system more flexible and suited for several human face editing applications. Finally, both quantitatively and qualitatively results proved that our system surpasses the current state of the art for semantic face generation and editing.

\clearpage  

%
%
\bibliographystyle{splncs04}
\bibliography{egbib}

\begin{thebibliography}{10}
\providecommand{\url}[1]{\texttt{#1}}
\providecommand{\urlprefix}{URL }
\providecommand{\doi}[1]{https://doi.org/#1}

\bibitem{baranchuk2021label}
Baranchuk, D., Rubachev, I., Voynov, A., Khrulkov, V., Babenko, A.: Label-efficient semantic segmentation with diffusion models. arXiv preprint arXiv:2112.03126  (2021)

\bibitem{croitoru2023diffusion}
Croitoru, F.A., Hondru, V., Ionescu, R.T., Shah, M.: Diffusion models in vision: A survey. IEEE Transactions on Pattern Analysis and Machine Intelligence  (2023)

\bibitem{dmbeatgan}
Dhariwal, P., Nichol, A.: Diffusion models beat gans on image synthesis. CoRR  \textbf{abs/2105.05233} (2021), \url{https://arxiv.org/abs/2105.05233}

\bibitem{fontanini2023frankenmask}
Fontanini, T., Ferrari, C., Lisanti, G., Galteri, L., Berretti, S., Bertozzi, M., Prati, A.: Frankenmask: Manipulating semantic masks with transformers for face parts editing. Pattern Recognition Letters  \textbf{176},  14--20 (2023)

\bibitem{textualInversion}
Gal, R., Alaluf, Y., Atzmon, Y., Patashnik, O., Bermano, A.H., Chechik, G., Cohen-Or, D.: An image is worth one word: Personalizing text-to-image generation using textual inversion (2022)

\bibitem{he2016deep}
He, K., Zhang, X., Ren, S., Sun, J.: Deep residual learning for image recognition. In: Proceedings of the IEEE conference on computer vision and pattern recognition. pp. 770--778 (2016)

\bibitem{heusel2017gans}
Heusel, M., Ramsauer, H., Unterthiner, T., Nessler, B., Hochreiter, S.: Gans trained by a two time-scale update rule converge to a local nash equilibrium. Advances in neural information processing systems  \textbf{30} (2017)

\bibitem{ho2020denoising}
Ho, J., Jain, A., Abbeel, P.: Denoising diffusion probabilistic models (2020)

\bibitem{dm}
Ho, J., Jain, A., Abbeel, P.: Denoising diffusion probabilistic models. CoRR  \textbf{abs/2006.11239} (2020), \url{https://arxiv.org/abs/2006.11239}

\bibitem{ho2021classifierfree}
Ho, J., Salimans, T.: Classifier-free diffusion guidance. In: NeurIPS 2021 Workshop on Deep Generative Models and Downstream Applications (2021), \url{https://openreview.net/forum?id=qw8AKxfYbI}

\bibitem{huang2023composer}
Huang, L., Chen, D., Liu, Y., Shen, Y., Zhao, D., Zhou, J.: Composer: Creative and controllable image synthesis with composable conditions (2023)

\bibitem{cdiffusion}
Huang, Z., Chan, K.C.K., Jiang, Y., Liu, Z.: Collaborative diffusion for multi-modal face generation and editing (2023)

\bibitem{karras2019stylebased}
Karras, T., Laine, S., Aila, T.: A style-based generator architecture for generative adversarial networks (2019)

\bibitem{celeba}
Lee, C.H., Liu, Z., Wu, L., Luo, P.: Maskgan: Towards diverse and interactive facial image manipulation (2020)

\bibitem{faceparsing}
liuziwei7: Celebamask-hq. \url{https://github.com/switchablenorms/CelebAMask-HQ/tree/master/face_parsing}

\bibitem{t2i}
Mou, C., Wang, X., Xie, L., Wu, Y., Zhang, J., Qi, Z., Shan, Y., Qie, X.: T2i-adapter: Learning adapters to dig out more controllable ability for text-to-image diffusion models. arXiv preprint arXiv:2302.08453  (2023)

\bibitem{spade}
Park, T., Liu, M.Y., Wang, T.C., Zhu, J.Y.: Semantic image synthesis with spatially-adaptive normalization (2019)

\bibitem{richardson2021encoding}
Richardson, E., Alaluf, Y., Patashnik, O., Nitzan, Y., Azar, Y., Shapiro, S., Cohen-Or, D.: Encoding in style: a stylegan encoder for image-to-image translation. In: Proceedings of the IEEE/CVF conference on computer vision and pattern recognition. pp. 2287--2296 (2021)

\bibitem{ldm}
Rombach, R., Blattmann, A., Lorenz, D., Esser, P., Ommer, B.: High-resolution image synthesis with latent diffusion models (2022)

\bibitem{ruiz2023dreambooth}
Ruiz, N., Li, Y., Jampani, V., Pritch, Y., Rubinstein, M., Aberman, K.: Dreambooth: Fine tuning text-to-image diffusion models for subject-driven generation (2023)

\bibitem{shi2022semanticstylegan}
Shi, Y., Yang, X., Wan, Y., Shen, X.: Semanticstylegan: Learning compositional generative priors for controllable image synthesis and editing (2022)

\bibitem{song2022denoising}
Song, J., Meng, C., Ermon, S.: Denoising diffusion implicit models (2022)

\bibitem{tan2021diverse}
Tan, Z., Chai, M., Chen, D., Liao, J., Chu, Q., Liu, B., Hua, G., Yu, N.: Diverse semantic image synthesis via probability distribution modeling. In: IEEE/CVF Conf. on Computer Vision and Pattern Recognition. pp. 7962--7971 (2021)

\bibitem{inade}
Tan, Z., Chai, M., Chen, D., Liao, J., Chu, Q., Liu, B., Hua, G., Yu, N.: Diverse semantic image synthesis via probability distribution modeling (2021)

\bibitem{clade}
Tan, Z., Chen, D., Chu, Q., Chai, M., Liao, J., He, M., Yuan, L., Hua, G., Yu, N.: Efficient semantic image synthesis via class-adaptive normalization (2021)

\bibitem{tarollo2024adversarial}
Tarollo, G., Fontanini, T., Ferrari, C., Borghi, G., Prati, A.: Adversarial identity injection for semantic face image synthesis. In: Proceedings of the IEEE/CVF Conference on Computer Vision and Pattern Recognition. pp. 1471--1480 (2024)

\bibitem{SDM}
Wang, W., Bao, J., Zhou, W., Chen, D., Chen, D., Yuan, L., Li, H.: Semantic image synthesis via diffusion models (2022)

\bibitem{cnet}
Zhang, L., Rao, A., Agrawala, M.: Adding conditional control to text-to-image diffusion models (2023)

\bibitem{zhu2020domain}
Zhu, J., Shen, Y., Zhao, D., Zhou, B.: In-domain gan inversion for real image editing. In: European conference on computer vision. pp. 592--608. Springer (2020)

\bibitem{sean}
Zhu, P., Abdal, R., Qin, Y., Wonka, P.: Sean: Image synthesis with semantic region-adaptive normalization. In: 2020 IEEE/CVF Conference on Computer Vision and Pattern Recognition (CVPR). IEEE (Jun 2020). \doi{10.1109/cvpr42600.2020.00515}, \url{http://dx.doi.org/10.1109/CVPR42600.2020.00515}

\end{thebibliography}

\newpage

\section*{Supplementary Material}

\begin{figure}[!h]
    \centering
    \includegraphics[width=\textwidth]{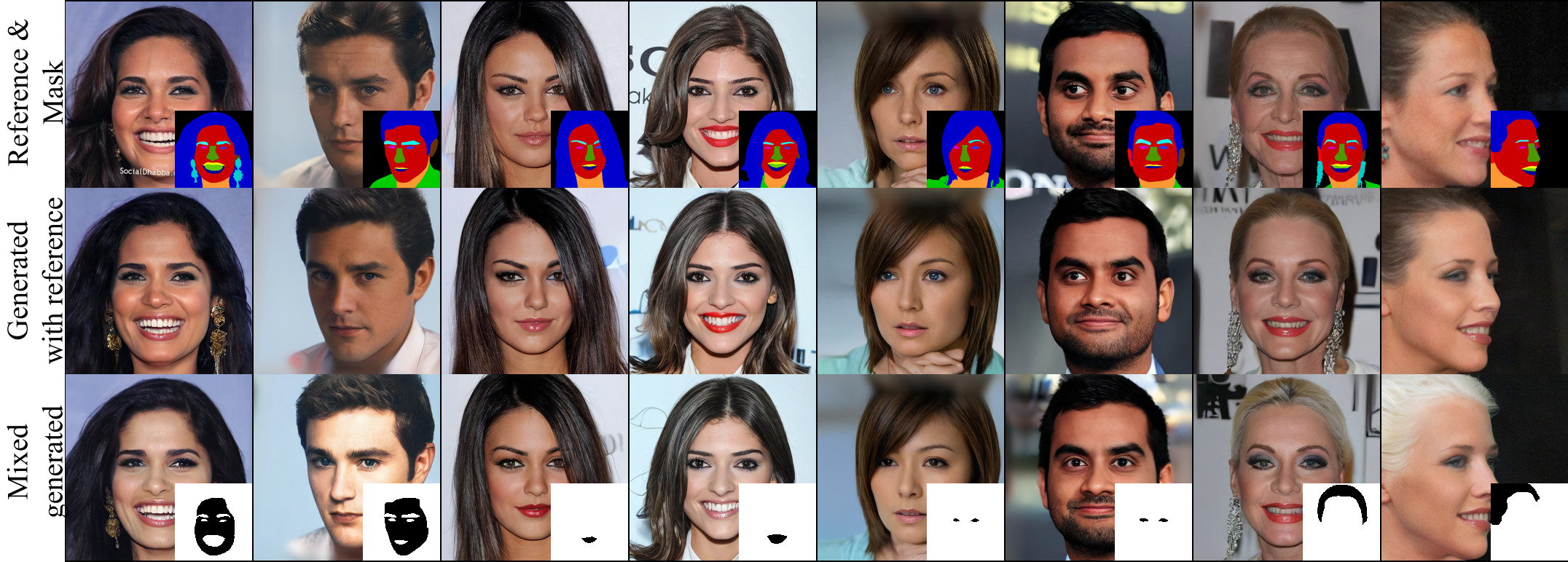}
    \caption{More examples of comparison between generating with full reference guidance and with some part randomly sampeld.}
    \label{fig:1}
\end{figure}
\begin{figure}[]
    \centering
    \includegraphics[width=\textwidth]{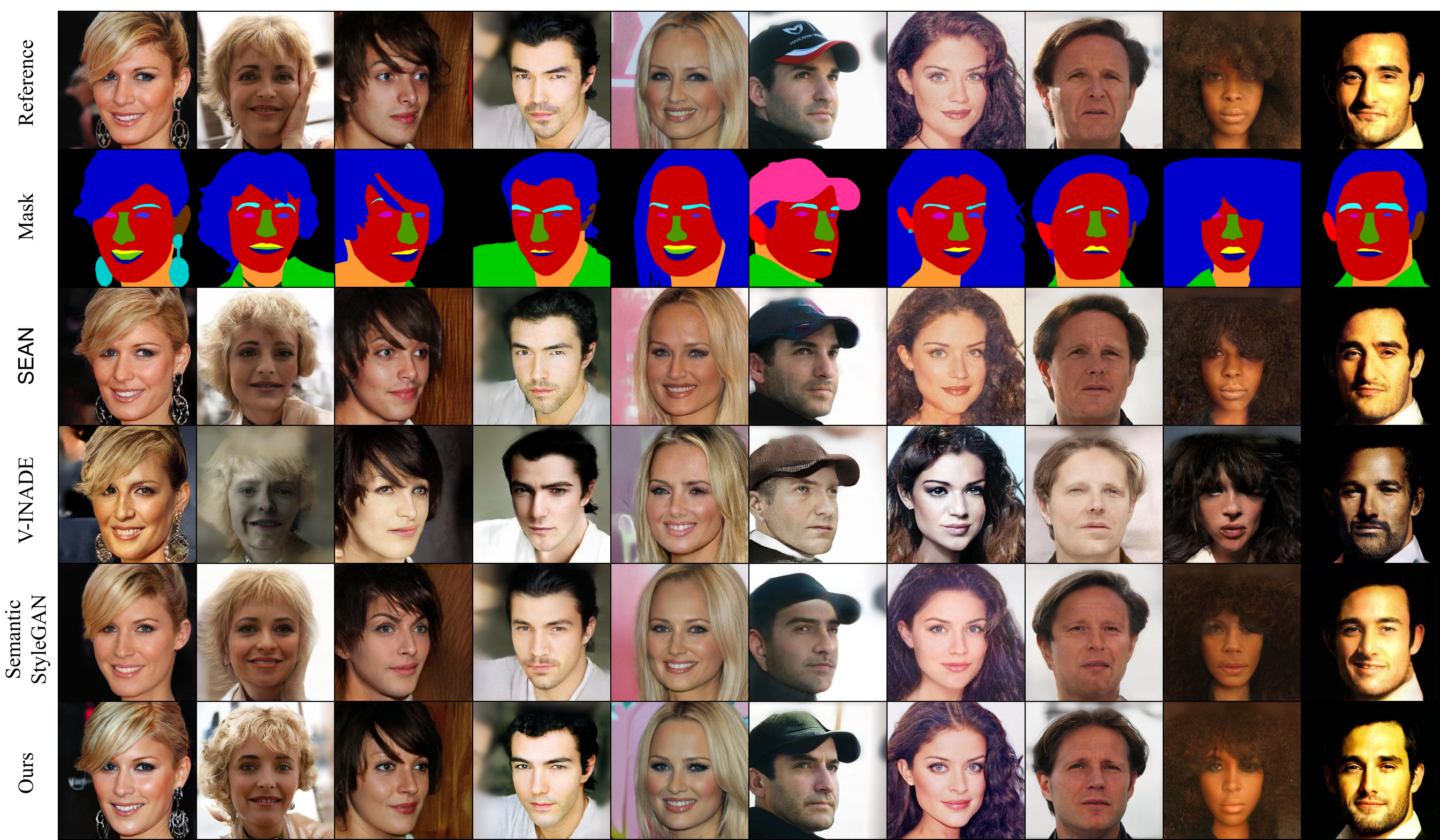}
    \caption{More examples of comparison between state-of-the-art models and ours.}
    \label{fig:2}
\end{figure}
\begin{figure}[]
    \centering
    \includegraphics[width=\textwidth]{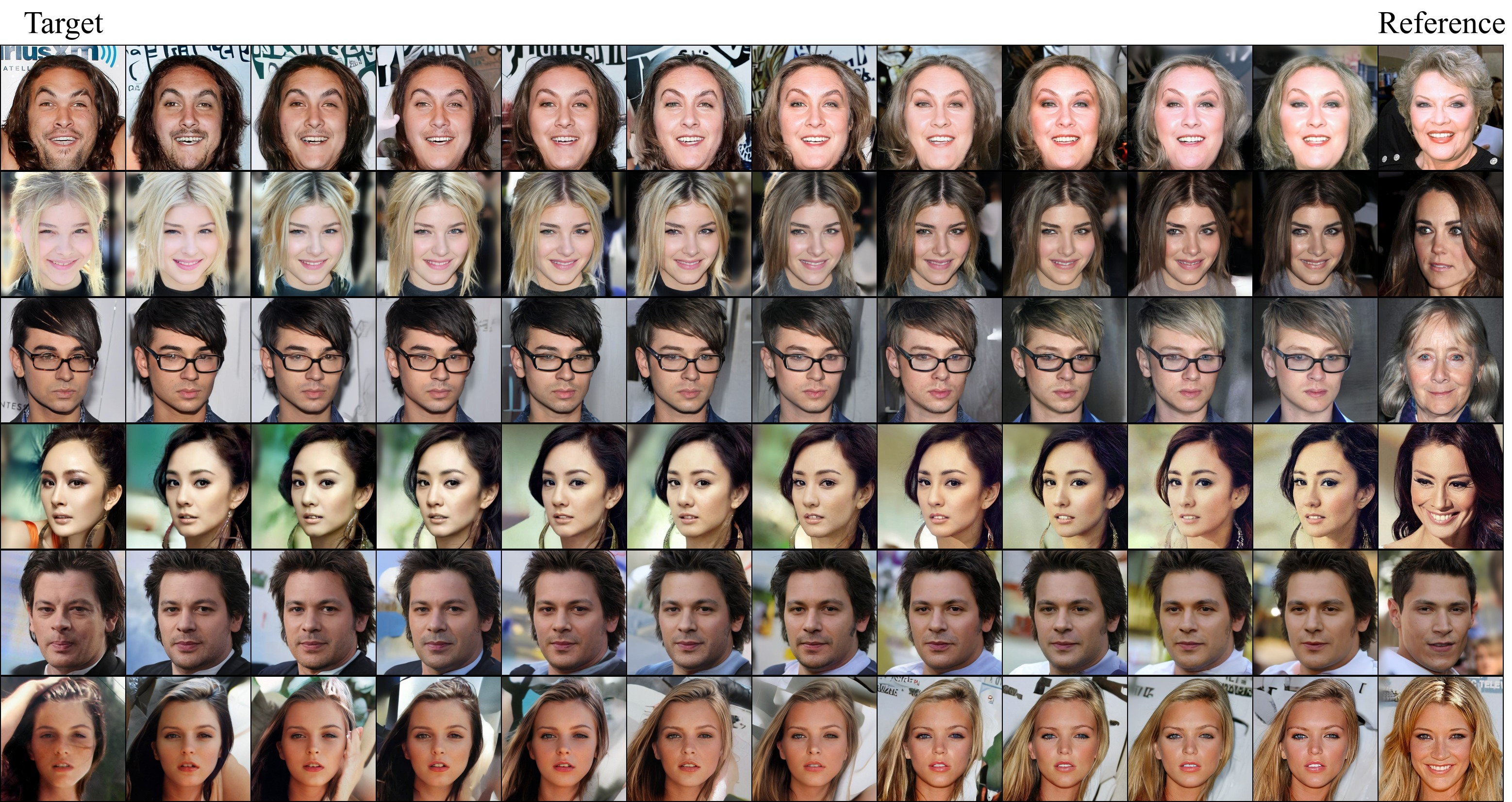}
    \caption{More examples of full interpolation from target (left) to full reference (right).}
    \label{fig:3}
\end{figure}
\begin{figure}[]
    \centering
    \includegraphics[width=\textwidth]{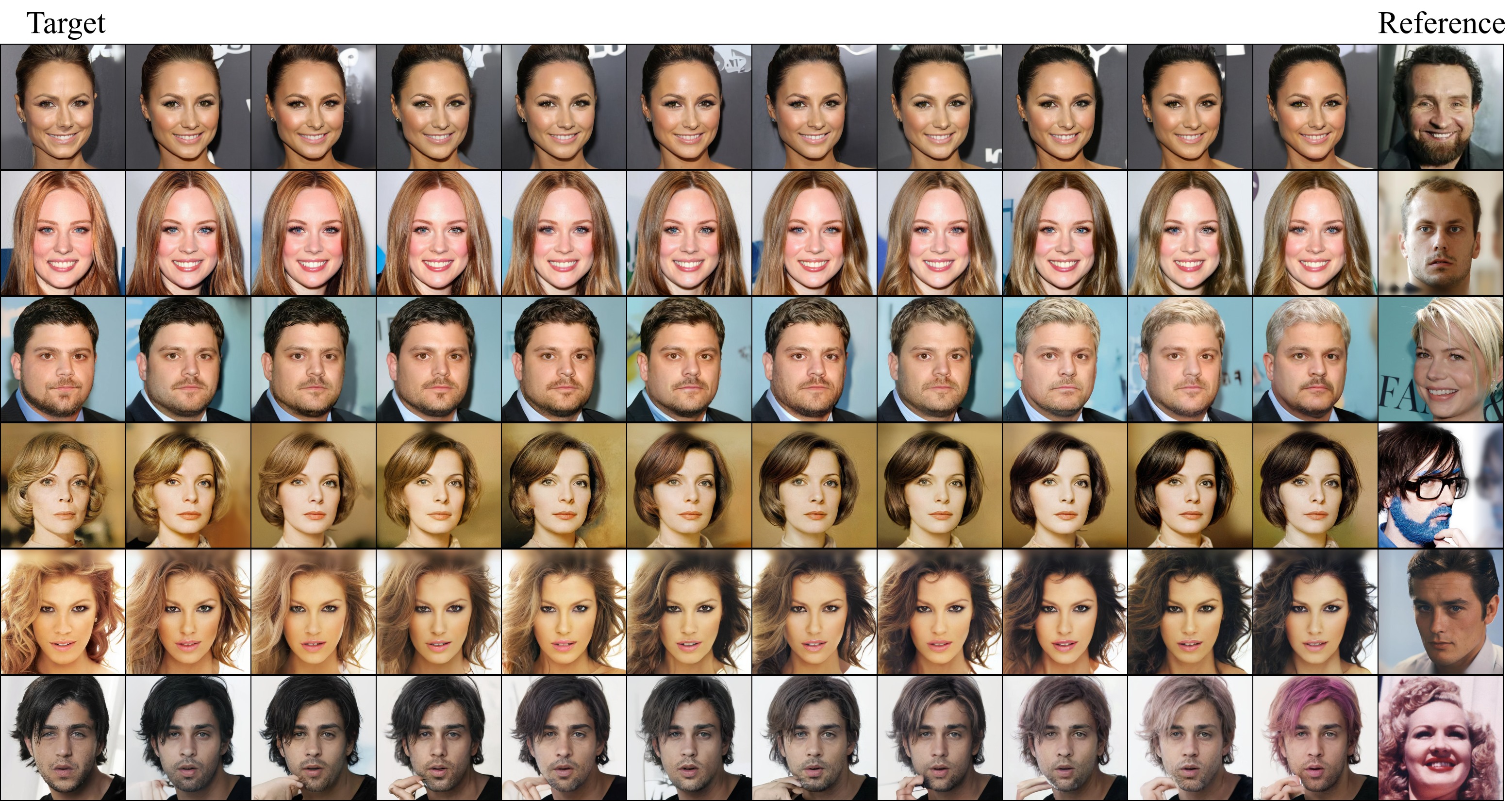}
    \caption{More examples of hair interpolation from full target (left) to full reference (right).}
    \label{fig:4}
\end{figure}
\begin{figure}[]
    \centering
    \includegraphics[width=\textwidth]{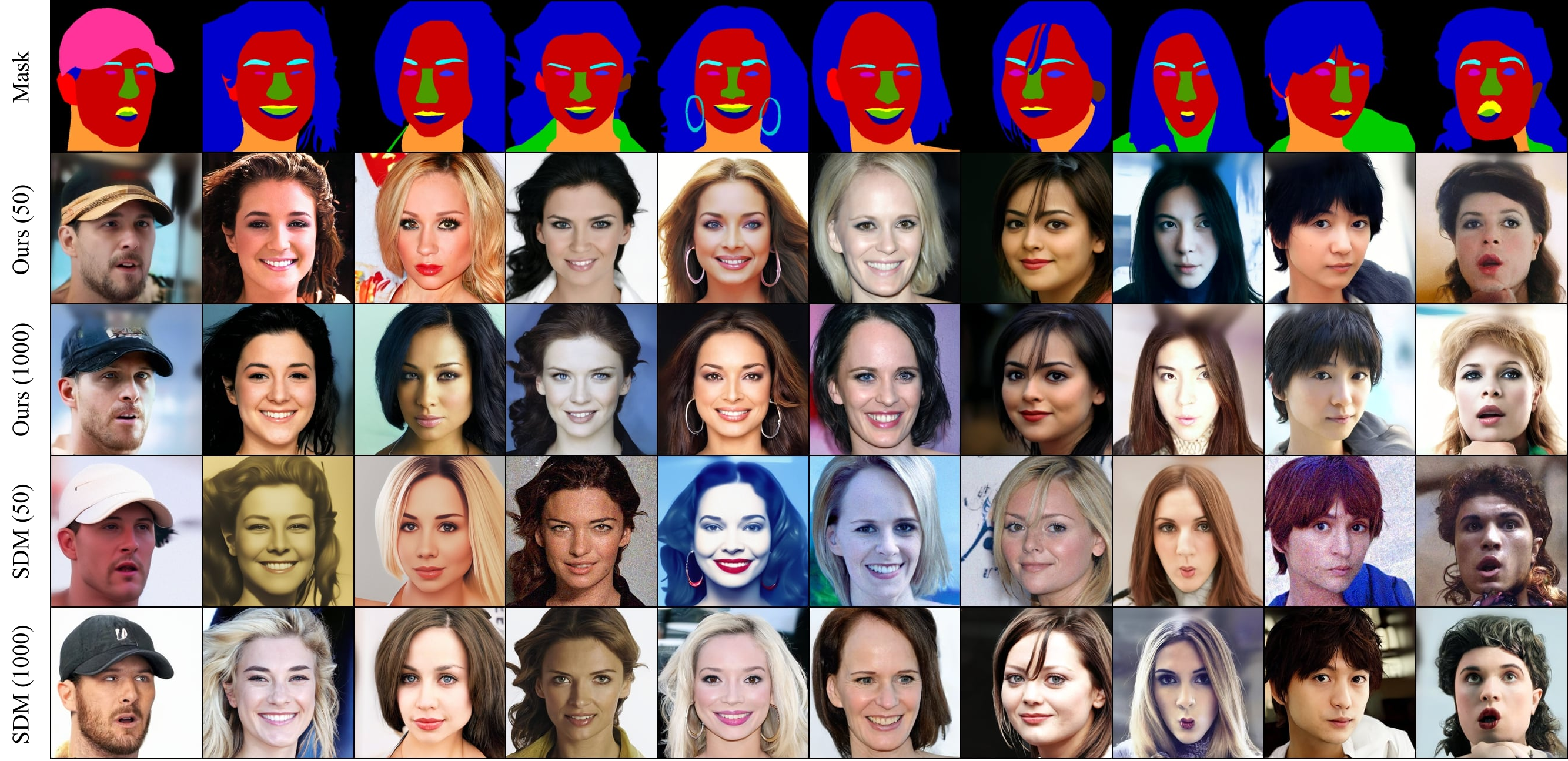}
    \caption{More examples of comparison between SDM and ours model at both 50 steps or 1000 steps.
    Inside the SDM samples there are some noisy ones, probably due to the fact that SDM works directly on pixel space, rather than on latent space, potentially resulting in a decline in the quality of generated samples. }
    \label{fig:5}
\end{figure}
\begin{figure}[]
    \centering
    \includegraphics[width=\textwidth]{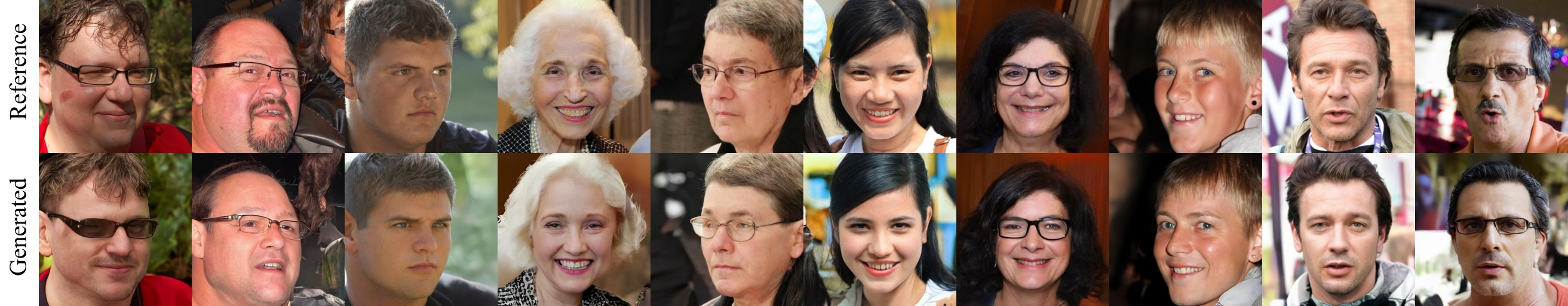}
    \caption{Qualitative outcomes illustrating the reconstruction performance of our model on FFHQ.}
    \label{fig:6}
\end{figure}

In this supplementary material we report additional qualitative examples. In particular, in Fig.~\ref{fig:1} other examples of mixed generation are reported to showcase the novel capability of our method to combine reference and noise-based generation. The examples are meant to highlight how the model can reconstruct a reference face image while simultaneously precisely generating the texture of specific face parts. In Fig.~\ref{fig:2} instead more examples of reference-based generation are reported in comparison with state of the art approaches. In Fig.~\ref{fig:3} and Fig.~\ref{fig:4}, other examples of style interpolation between two reference and target images are shown, both for local and global styles. These highlight the disentanglement ability as provided by the masked cross-attention layers. In Fig.~\ref{fig:5} we report additional qualitative comparisons between our method and Semantic Diffusion Model (SDM)~\cite{SDM}. This image is meant to highlight that our model performs qualitatively on par when using 1000 diffusion steps, but provided more stable and accurate results if reducing the number of steps so to decrease the time required for generating the images. Finally, in Fig.~\ref{fig:6} we present a qualitative capacity demonstrated by our model, which was trained using CelebA-HQ~\cite{celeba} data but evaluated on FFHQ ~\cite{karras2019stylebased} data. This further proves the flexibility of the proposed system.

\end{document}